\documentclass[11pt]{article}

\usepackage[final]{acl}

\usepackage{times}
\usepackage{latexsym}

\usepackage[T1]{fontenc}

\usepackage[utf8]{inputenc}

\usepackage{microtype}

\usepackage{inconsolata}

\usepackage{graphicx}
\usepackage{multirow}
\usepackage{arydshln}

\title{\textsc{LexRubric}: A Rubric-Guided Diagnostic Benchmark for Open-Ended Legal Tasks}

\author{
  \textbf{Yifan Chen\textsuperscript{1}\thanks{These authors contributed equally to this work.}},
  \textbf{Haitao Li\textsuperscript{1}\footnotemark[1]},
  \textbf{Yiran Hu\textsuperscript{2}},
  \textbf{Kaisong Song\textsuperscript{3}},
  \textbf{Jun Lin\textsuperscript{3}},
\\
  \textbf{Yueyue Wu\textsuperscript{4}\thanks{Corresponding authors.}},
  \textbf{Qingyao Ai\textsuperscript{1}\footnotemark[2]},
  \textbf{Min Zhang\textsuperscript{1}},
  \textbf{Yiqun Liu\textsuperscript{1}}
\\
  \textsuperscript{1}Department of Computer Science and Technology, Tsinghua University,
\\
  \textsuperscript{2}University of Waterloo,
  \textsuperscript{3}Alibaba Group,
  \textsuperscript{4}Tsinghua University Think Tank Center
\\
  \texttt{chenyifan26@mails.tsinghua.edu.cn}
}

\begin{document}
\maketitle
\begin{abstract}
As large language models (LLMs) are increasingly applied to real-world legal tasks, evaluating the reliability of their open-ended legal responses has become essential. These tasks require context-sensitive answers and allow little room for error, motivating fine-grained and diagnostic evaluation that can identify specific sources of response quality failures. We introduce \textsc{LexRubric}, a rubric-based benchmark for evaluating open-ended Chinese legal tasks. \textsc{LexRubric} contains 649 instances from legal consultation and judicial examination, which reflect both everyday legal needs and professional legal reasoning and cover 14 legal scenarios. It further includes 12,337 expert-written atomic scoring criteria organized under a unified six-dimensional framework, enabling accurate evaluation and diagnostic analysis across tasks and evaluation dimensions. To validate the reliability of the evaluation, we test multiple judge models and compare model-based judgments with human judgments. We further evaluate 18 recent general and legal-domain LLMs on \textsc{LexRubric}. Results show that different models exhibit distinct capability profiles, and that open-ended legal tasks remain challenging for current LLMs. Data is available at: \url{https://github.com/foggpoy/LexRubric}.
\end{abstract}

\section{Introduction}

Large language models (LLMs) have rapidly improved in natural language understanding and generation, and are increasingly being applied in the legal domain \citep{brown2020language,openai2023gpt4,lai2024large,dehghani2025legalsystems,li2026legalonefamilyfoundationmodels}. In real-world legal tasks, both lay users and legal professionals often raise open-ended and context-dependent questions. Reliable responses to such open-ended questions must coordinate multiple aspects, including applicable rules, factual context, legal reasoning, and practical implications, across diverse possible contents and structures. At the same time, legal-domain responses require every substantive part to be accurate and non-misleading \citep{magesh2025hallucination,hu2026evaluation}. These properties require fine-grained and multi-dimensional evaluation of open-ended legal tasks, supporting both comprehensive assessment and diagnostic analysis.

Most existing legal benchmarks are still designed around standardized task formulations or standardized evaluation methods, limiting their ability to reflect the usability of model outputs in real-world legal question answering. Early benchmarks such as CAIL2018 \citep{xiao2018cail2018} and CUAD \citep{hendrycks2021cuad} mainly evaluate specific legal tasks through predefined labels, answer spans, or task-specific metrics. Comprehensive benchmarks, including LegalBench \citep{guha2023legalbench}, LawBench \citep{fei2024lawbench}, LAiW \citep{dai2025laiw}, and LexEval \citep{li2024lexeval}, further expand the coverage of legal knowledge, reasoning, and application abilities. However, their evaluation methods still largely rely on standardized answers, task-specific metrics, or aggregate performance scores. These designs support reproducible comparison, but provide limited evidence for accurate assessment and diagnostic analysis in open-ended legal tasks.

\begin{figure*}[t]
  \centering
  \includegraphics[width=\textwidth]{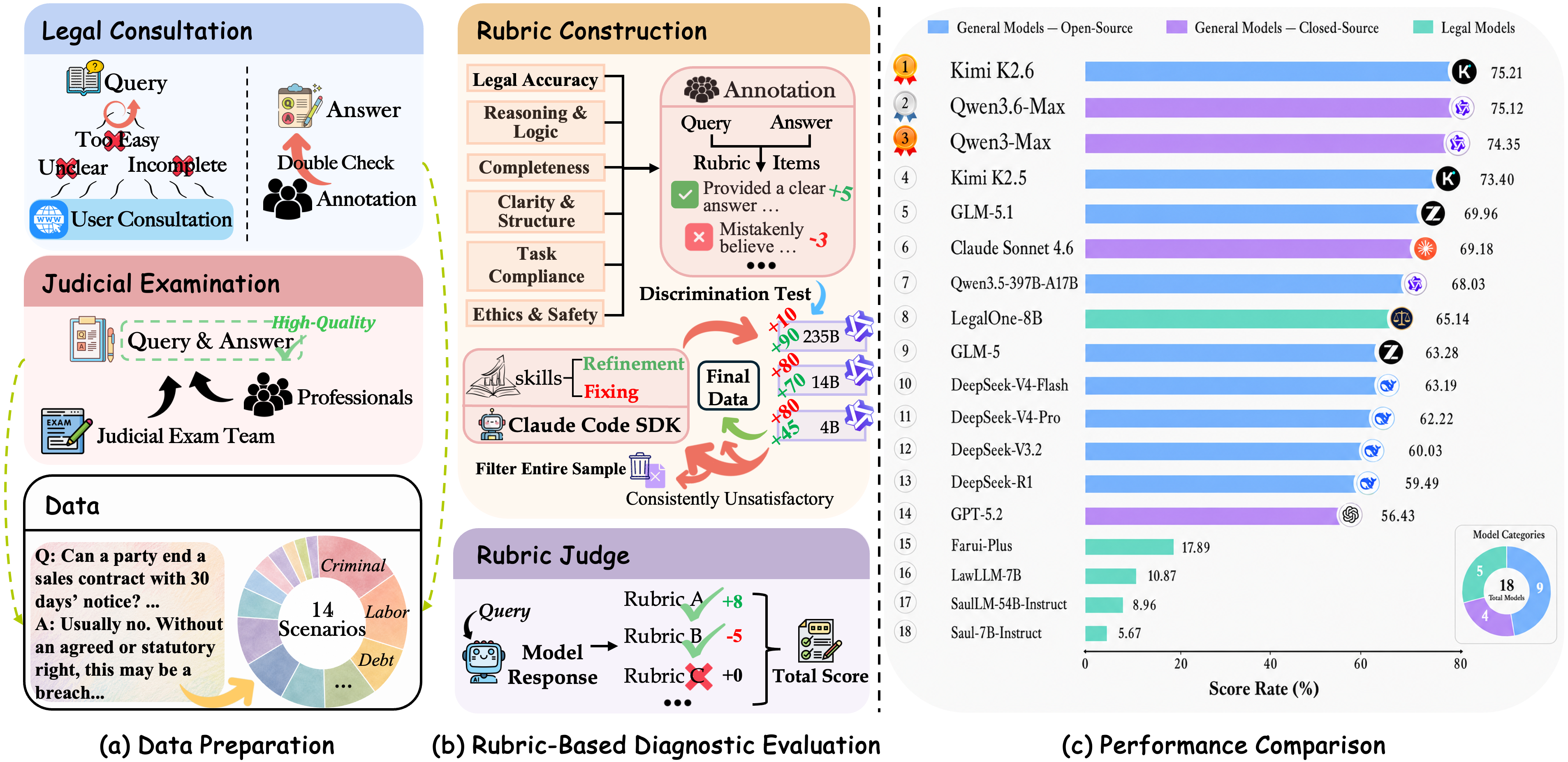}
  \caption{Overview framework of \textsc{LexRubric}: the left side shows the construction and evaluation workflow, while the right side compares model performance on the benchmark.}
  \label{fig:overview}
\end{figure*}

Practice-oriented legal benchmarks further improve task realism, but fine-grained diagnostic evaluation of open-ended legal tasks remains underdeveloped. UCL-Bench adopts a user-centric design based on legal practitioners' needs, but its evaluation is mainly organized around task fulfillment and answer guidance; its relatively coarse and weakly differentiated evaluation hints limit fine-grained assessment \citep{gan2025uclbench}. PLawBench is closely aligned with professional legal workflows, but its evaluation uses task-specific and relatively composite criteria for different legal task types \citep{shi2026plawbench}. These designs motivate a benchmark that centers on accurate evaluation of open-ended legal tasks and presents model performance in an interpretable way.

We introduce \textsc{LexRubric}, a rubric-based benchmark for evaluating open-ended Chinese legal tasks (Figure~\ref{fig:overview}). Its core idea is to combine expert judgment for each legal question with a unified diagnostic evaluation framework, so that evaluation can better reflect human expert judgments while also supporting cross-task comparison. The benchmark covers two complementary sources of legal questions: \emph{legal consultation}, which is derived from real user queries and reflects diverse practical needs, and \emph{judicial examination}, which provides professionally designed and knowledge-intensive questions. Inspired by rubric-based evaluation in high-stakes professional domains \citep{arora2025healthbench,akyurek2025prbench,gunjal2025rubricsrewardsreinforcementlearning}, \textsc{LexRubric} uses expert-written scoring criteria to evaluate each instance. These criteria are decomposed into fine-grained assessment units and organized under six shared dimensions. By combining instance-specific expert assessment with a consistent evaluation framework, \textsc{LexRubric} supports precise evaluation of individual responses and diagnostic comparison across tasks, scenarios, and quality dimensions.

Our contributions are threefold:
\begin{itemize}
    \item \textbf{Open-ended Chinese legal benchmark.} We construct \textsc{LexRubric}, covering 649 instances from legal consultation and judicial examination across 14 legal scenarios. The benchmark targets naturally formulated legal questions that require open-ended responses, rather than closed-form selection, classification, or generation under a fixed answer structure.
    \item \textbf{Fine-grained and diagnostic rubric-based evaluation framework.} We design a unified six-dimensional quality framework and instantiate it with 12,337 expert-written scoring criteria. This enables precise response-level assessment and supports item-level, dimension-level, and task-level analysis of model behavior.
    \item \textbf{Systematic analysis of LLMs' legal capabilities.} We evaluate both general-purpose and legal-domain LLMs on \textsc{LexRubric}. The results show that the benchmark reveals model strengths and weaknesses across tasks, dimensions, and different models, including patterns that are difficult to observe from a single aggregate score.
\end{itemize}

\section{Related Work}

\begin{figure*}[t]
  \centering
  \includegraphics[width=\textwidth]{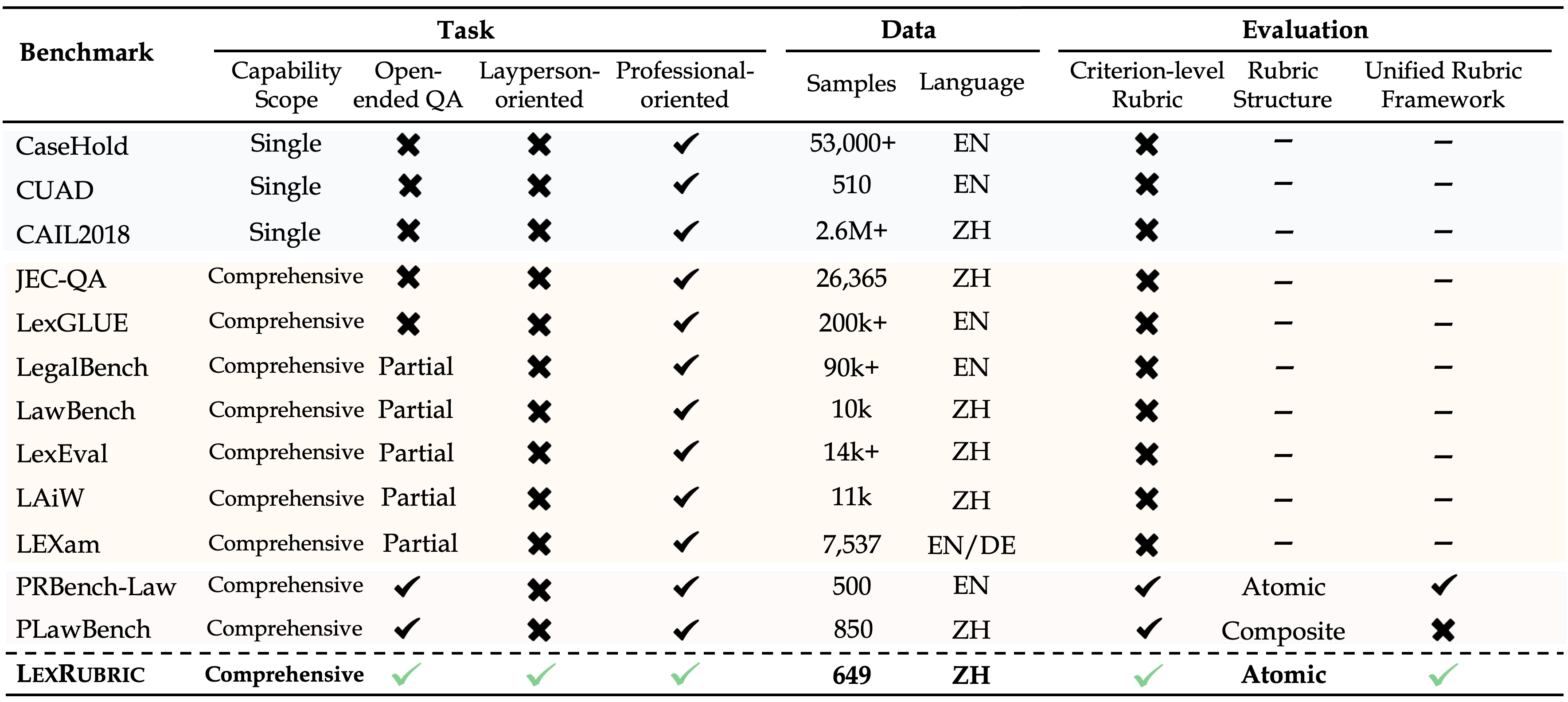}
  \caption{Comparison between \textsc{LexRubric} and representative legal benchmarks. \emph{Capability scope} indicates the breadth of abilities evaluated; \textit{single} refers to a specific capability (e.g., judgment prediction or contract review). \emph{Layperson-oriented} denotes legal questions that ordinary users may encounter in daily life; in \textsc{LexRubric}, such queries are collected from real user queries. \emph{Professional-oriented} denotes questions mainly arising in professional legal settings, such as legal knowledge, legal reasoning, or exam-style tasks.}
  \label{fig:benchmark_comparison}
\end{figure*}

As artificial intelligence (AI) systems are increasingly applied to legal reasoning and decision-making tasks, evaluating model capabilities has become an important foundation of legal AI research \citep{lai2024large,dehghani2025legalsystems,hu2026evaluation}. Early legal benchmarks mainly focused on specific legal tasks or technical settings, covering areas such as judgment prediction, contract review, legal retrieval-augmented generation, and legal agent evaluation \citep{xiao2018cail2018,hendrycks2021cuad,yao2022leven,li2023lecardv2,pipitone2024legalbenchrag,li2025lexrag,li2025legalagentbench,li2025casegen}. These works make targeted legal abilities measurable. Yet a benchmark tied to a single task setting has limited capacity to reflect a model's overall performance in practical legal use.

Legal benchmarks have further extended this line from task-specific evaluation to broader capability coverage. LexGLUE~\citep{chalkidis2022lexglue}, LegalBench~\citep{guha2023legalbench}, LawBench~\citep{fei2024lawbench}, LAiW~\citep{dai2025laiw}, LexEval~\citep{li2024lexeval}, and DISC-Law-Eval~\citep{yue2023disclawllm} organize legal evaluation across knowledge, reasoning, application, and system-level abilities. These benchmarks define the basic landscape of legal LLM evaluation. Their emphasis is mainly task coverage, ability taxonomy, and standardized scoring, often through answer options, fixed outputs, or short references \citep{zhong2020jecqa,zheng2021casehold,fan2026lexambenchmarkinglegalreasoning}. \textsc{LexRubric} instead focuses on open-ended complex legal question answering.

Recent benchmarks have moved closer to realistic legal use. UCL-Bench~\citep{gan2025uclbench} improves Chinese legal evaluation through user-centered legal needs and professional answer guidance, although its feedback remains relatively coarse. Rubric-based evaluation offers a structured way to assess open-ended professional-domain question answering \citep{arora2025healthbench,akyurek2025prbench}. Its use as training signals further suggests that rubrics can provide actionable feedback beyond final scores \citep{gunjal2025rubricsrewardsreinforcementlearning}.

PLawBench~\citep{shi2026plawbench} brings rubric-based evaluation into legal practice. It covers lawyer-client consultation, case analysis, and document generation, and designs task-specific rubric structures for different professional workflows. However, its rubric items can be relatively composite, requiring a judge to parse multiple assessment requirements within one item. In contrast, \textsc{LexRubric} targets broader open-ended legal LLM applications, including legal problems that professionals and ordinary users may encounter. It uses a shared six-dimensional framework across tasks and adopts point-level atomic criteria within each instance. The atomic design reduces the parsing burden for cost-effective LLM judges and makes rubric checking more accurate and comparable. Figure~\ref{fig:benchmark_comparison} summarizes the differences between \textsc{LexRubric} and representative legal benchmarks.

\section{\textsc{LexRubric}}
\label{sec:lexrubric}

\textsc{LexRubric} is designed to evaluate open-ended Chinese legal tasks in a way that is close to real use, comparable across tasks, and useful for diagnosis. The benchmark consists of legal questions, expert-written reference answers, and instance-specific atomic rubrics for fine-grained assessment of model responses. Figure~\ref{fig:overview} shows the overall data construction and evaluation workflow.

\subsection{Design}
\label{sec:design}

The design of \textsc{LexRubric} follows five goals. \emph{Realistic}: the tasks should reflect real Chinese legal use scenarios. \emph{Open-ended}: model outputs should be open-ended, rather than limited to answer options or other fixed forms and structures. \emph{Diagnostic}: the evaluation should reveal different models' strengths and weaknesses. \emph{Comparable}: different task types should be analyzed under a shared framework. \emph{Reliable}: all reference answers and scoring criteria are annotated by legal experts and further checked through a quality-control pipeline.

We construct \textsc{LexRubric} from two complementary sources. The first source is legal consultation. We collect real-world user queries concerning legal issues, spanning diverse user roles including parties, legal practitioners, companies, and public institutions. These questions cover practical needs such as legal judgment, risk assessment, dispute strategy, document assistance, and compliance advice. The second source is judicial examination. Legal experts write examination-style questions in collaboration with a judicial-examination platform. These questions contain denser legal knowledge and more explicit normative analysis. Together, the two sources cover a spectrum from everyday legal needs to professional legal capabilities.

\subsection{Data Collection}
\label{sec:data_collection}

For legal consultation, we start from more than 50,000 real user queries. We score these queries by difficulty, completeness, practical value, legal relevance, and answerability for filtering. Queries with unclear expression but valuable legal intent are rewritten to improve clarity while preserving the original legal problem. After filtering, 622 consultation questions are selected for expert annotation. The consultation split also includes a supplementary subset of 40 Chinese law-related items from OneMillion-Bench~\citep{yang2026onemillionbenchfarlanguageagents}, for which we retain the original rubrics and annotate reference answers.

For judicial examination, legal experts write 250 questions for annotation. These questions are designed to cover professional legal reasoning and knowledge-intensive analysis.

For all selected questions, experts first write reference answers and then construct rubrics. The reference answers guide rubric construction, but the final evaluation does not rely on exact matching to a single gold answer.

\subsection{Rubric Construction}
\label{sec:rubric_construction}

For each instance, legal experts construct a dedicated set of rubric items. Each rubric item specifies a concrete requirement that a response should satisfy or avoid, together with a point value from \(-10\) to \(10\). Positive items describe desirable qualities of a high-quality response, while negative items describe undesirable, incorrect, unsafe, or misleading properties. The absolute value of the point reflects the relative importance of the requirement. Since instances differ in difficulty and complexity, the number of rubric items and the total possible score vary across instances.

Before instance-level rubric construction, legal experts first developed a set of consensus standards. These standards define objective levels for recurring procedural requirements in legal responses and specify the scoring expectations for each level, ensuring consistency and rigor in subsequent expert annotation. The standards cover eight common categories: \emph{emergency legal procedure guidance}, \emph{information seeking}, \emph{cross-jurisdiction adaptation}, \emph{legal document handling}, \emph{communication customization}, \emph{responses under uncertainty}, \emph{response depth and legal reasoning}, and \emph{ethics and safety}. These standards define the appropriate response requirements for common situations, such as missing legal context, jurisdiction-dependent answers, urgent procedural risks, user-role differences, legal uncertainty, and potentially unsafe or abusive requests. Detailed descriptions of the consensus standards are provided in Appendix~\ref{app:consensus_standards}.

The instance-level rubrics are then constructed by multiple legal experts. Experts write both general criteria derived from the consensus standards and question-specific criteria tailored to the legal facts, issues, and expected reasoning of each instance. During rubric construction, experts follow six principles:

\begin{itemize}
    \item \textbf{Valid.} Each criterion must be legally correct and unambiguous.

    \item \textbf{Task relevance.} Each criterion should be grounded in the given question and the expected legal task, without introducing irrelevant or excessive requirements.
    
    \item \textbf{Mutually exclusive and relatively complete.} Criteria within the same rubric set should not repeatedly assess the same legal point. Besides, the rubrics should cover the key aspects of an ideal answer and avoid omitting core requirements.
    
    \item \textbf{Atomic.} Each criterion should assess only one requirement. For example, citing a legal provision, identifying applicable conditions, and analyzing factual application should not be bundled into a single criterion.
    
    \item \textbf{Objective and binary.} Each criterion should be formulated so that the judgment result is limited to either satisfied or not satisfied. It should not require annotators to evaluate the degree or extent to which the response satisfies the requirement.
    
    \item \textbf{Self-contained.} Each criterion should be assessable from the model response itself, without requiring reference to other criteria or external materials.
\end{itemize}

\textsc{LexRubric} organizes all rubric items under six dimensions. These dimensions separate legal substance from general response quality, while remaining general enough for consultation, examination-style reasoning, and practical analysis. Table~\ref{tab:six_dimensions} summarizes the framework.

\begin{table}[t]
  \centering
  {
  \small
  \begin{tabular}{p{0.35\columnwidth}p{0.54\columnwidth}}
    \hline
    \textbf{Dimension} & \textbf{Main focus} \\
    \hline
    Legal Accuracy & Correct use of legal rules, legal relations, liability, penalties, and uncertainty. \\
    Reasoning \& Logic & Clear and coherent legal reasoning supported by facts and rules. \\
    Completeness & Coverage of key facts, legal elements, procedures, risks, and exceptions. \\
    Clarity \& Structure & Clear organization, appropriate wording, and user-matched technical depth. \\
    Task Compliance & Relevance to the question, instruction following, and required format. \\
    Ethics \& Safety & Avoidance of unsafe, misleading, unlawful, or overconfident advice. \\
    \hline
  \end{tabular}
  }
  \caption{Six-dimensional evaluation framework in \textsc{LexRubric}.}
  \label{tab:six_dimensions}
\end{table}

Formally, for an instance \(x_i\), its rubric set is
\(\mathcal{R}_i=\{(c_{ij}, p_{ij}, d_{ij})\}_{j=1}^{m_i}\), where \(c_{ij}\) is the criterion, \(p_{ij}\in[-10,10]\) is the point value, and \(d_{ij}\) is the dimension. Given a model response \(y_i\), its score is:
\begin{equation}
  \label{eq:rubric_score}
  S(x_i,y_i)=\sum_{j=1}^{m_i} p_{ij}\cdot \mathbf{1}\{y_i \models c_{ij}\},
\end{equation}
where \(\mathbf{1}\{y_i \models c_{ij}\}\) indicates whether the response triggers the criterion.

\paragraph{Annotation cost.}
All annotators are legal practitioners or Ph.D. students from top universities who have passed the National Unified Legal Profession Qualification Examination. Each instance undergoes three independent annotation rounds, each completed by a different legal expert. The compensation for annotating the reference answer and rubric set is \$44.1--\$73.5 per expert per instance, depending on difficulty. In total, 872 instances are annotated, resulting in an overall annotation cost of approximately \$154,000.

\subsection{Rubric Refinement}
\label{sec:rubric_refinement}

\begin{table}[t]
  \centering
  {
  \small
  \begin{tabular}{lrrr}
    \hline
    \textbf{Statistic} & \textbf{Consult.} & \textbf{Exam} & \textbf{Total} \\
    \hline
    Number of instances & 473 & 176 & \textbf{649} \\
    Avg. question length & 818.82 & 1015.15 & \textbf{872.06} \\
    Number of rubric items & 10,601 & 1,736 & \textbf{12,337} \\
    Avg. rubric items & 22.41 & 9.86 & \textbf{19.01} \\
    Avg. rubric length & 55.98 & 45.90 & \textbf{54.56} \\
    \hline
  \end{tabular}
  }
  \caption{Dataset statistics of \textsc{LexRubric}.}
  \label{tab:data_statistics}
\end{table}

\begin{table}[t]
  \centering
  {
  \small
  \begin{tabular}{lrrr}
    \hline
    \textbf{Dimension} & \textbf{Consult.} & \textbf{Exam} & \textbf{Total} \\
    \hline
    Legal Accuracy & 3,318 & 629 & \textbf{3,947} \\
    Reasoning \& Logic & 3,426 & 725 & \textbf{4,151} \\
    Completeness & 1,425 & 80 & \textbf{1,505} \\
    Clarity \& Structure & 485 & 108 & \textbf{593} \\
    Task Compliance & 1,458 & 183 & \textbf{1,641} \\
    Ethics \& Safety & 489 & 11 & \textbf{500} \\
    \hline
    Total & 10,601 & 1,736 & \textbf{12,337} \\
    \hline
  \end{tabular}
  }
  \caption{Distribution of rubric items across dimensions.}
  \label{tab:dimension_statistics}
\end{table}

To ensure rubric quality, we adopt an expert-in-the-loop construction process. For each instance, multiple legal experts independently construct candidate rubrics, which are then manually reviewed and consolidated into a final rubric set. This process mitigates the limitations of individual expert judgment.

We further apply a quality-control pipeline to improve rubric discriminativeness and validity. We use three Qwen3 models with distinguishable capability levels: Qwen3-235B-A22B, Qwen3-14B, and Qwen3-4B \citep{yang2025qwen3technicalreport}. Using models from the same family reduces architectural and training variation, while model scale provides a coarse capability ordering. We use this ordering to flag suspicious cases with reversed score gradients or weak discriminativeness. For each instance, we generate model responses and score them with the annotated rubrics. Instances are flagged when the scores clearly violate the expected capability gradient or when the rubrics fail to distinguish responses of different quality. For flagged instances, an AI-assisted workflow based on Claude Code SDK\footnote{\url{https://code.claude.com/docs/en/agent-sdk}} and legal-domain skills, such as a legal research skill\footnote{\url{https://github.com/Golden2002/legal-research-skill}}, is used to identify and repair rubric defects, including vagueness, overbreadth, or weak discriminativeness. The workflow separates rubric defects from model errors, and all AI-assisted revisions are reviewed and confirmed by legal experts. Instances that remain weakly discriminative after refinement are filtered out.

Finally, \textsc{LexRubric} contains 649 instances across 14 legal scenarios. Table~\ref{tab:data_statistics} reports dataset statistics, and Table~\ref{tab:dimension_statistics} reports the distribution of rubric items across dimensions. We provide the detailed scenario distribution and concrete data examples in Appendix~\ref{sec:dataset_details}.

\section{Experiments}
\label{sec:experiments}

\begin{table*}[t]
  \centering
  {
  \small
  \setlength{\tabcolsep}{3.8pt}
  \begin{tabular*}{\textwidth}{@{\extracolsep{\fill}}lrrr|rrrrrr}
    \hline
    \multirow{2}{*}{\textbf{Model}} &
    \multicolumn{3}{c|}{\textbf{Overall Score Rate (\%)}} &
    \multicolumn{6}{c}{\textbf{Dimension Score Rate (\%)}} \\
    \cline{2-4}\cline{5-10}
    & \textbf{Consult.} & \textbf{Exam} & \textbf{All}
    & \textbf{Acc.} & \textbf{R\&L} & \textbf{T.Comp.} & \textbf{E\&S} & \textbf{Compl.} & \textbf{C\&S} \\
    \hline
    Kimi K2.6 & \underline{72.84} & \textbf{81.59} & \textbf{75.21}
    & \underline{68.53} & \textbf{79.08} & \underline{86.97} & 42.55 & \textbf{73.85} & 92.74 \\
    Qwen3.6-Max-Preview & \textbf{73.17} & 80.35 & \underline{75.12}
    & \textbf{69.51} & \underline{77.14} & \textbf{87.30} & 59.82 & \underline{72.02} & \textbf{94.62} \\
    Qwen3-Max & 71.88 & \underline{80.99} & 74.35
    & 68.45 & 77.07 & 84.06 & \underline{69.64} & 70.84 & \underline{92.90} \\
    Kimi K2.5 & 70.79 & 80.39 & 73.40
    & 66.60 & 76.50 & 86.94 & 40.86 & 70.91 & 92.41 \\
    GLM-5.1 & 67.57 & 76.37 & 69.96
    & 62.13 & 74.72 & 82.83 & 26.84 & 65.15 & \underline{92.90} \\
    Claude Sonnet 4.6 & 70.09 & 66.73 & 69.18
    & 60.60 & 72.24 & 81.91 & \textbf{72.74} & 69.71 & 92.37 \\
    Qwen3.5-397B-A17B & 65.85 & 73.88 & 68.03
    & 59.46 & 72.17 & 82.32 & 56.67 & 64.95 & 91.36 \\
    LegalOne-8B & 62.93 & 71.10 & 65.14
    & 61.65 & 67.31 & 78.34 & 53.31 & 59.18 & 89.46 \\
    GLM-5 & 60.02 & 72.04 & 63.28
    & 56.92 & 68.24 & 77.01 & 22.03 & 54.98 & 90.45 \\
    DeepSeek-V4-Flash & 62.64 & 64.67 & 63.19
    & 55.34 & 65.71 & 80.01 & 54.72 & 61.11 & 88.21 \\
    DeepSeek-V4-Pro & 58.24 & 72.93 & 62.22
    & 54.21 & 65.86 & 78.20 & 27.59 & 54.10 & 87.47 \\
    DeepSeek-V3.2 & 55.96 & 70.96 & 60.03
    & 54.17 & 61.36 & 75.35 & 61.16 & 52.93 & 88.95 \\
    DeepSeek-R1 & 58.23 & 62.86 & 59.49
    & 52.97 & 62.43 & 75.49 & 40.64 & 57.20 & 88.33 \\
    GPT-5.2 & 57.22 & 54.28 & 56.43
    & 37.75 & 63.87 & 77.68 & 34.70 & 63.94 & 85.99 \\
    Farui-Plus & 15.61 & 24.03 & 17.89
    & 16.03 & 14.60 & 34.40 & 24.46 & 13.91 & 50.77 \\
    LawLLM-7B & 9.55 & 14.40 & 10.87
    & 8.25 & 7.78 & 25.54 & 18.61 & 5.45 & 33.71 \\
    SaulLM-54B-Instruct & 9.05 & 8.73 & 8.96
    & 3.11 & 7.77 & 26.86 & 8.51 & 10.32 & 28.30 \\
    Saul-7B-Instruct & 5.93 & 4.96 & 5.67
    & 1.30 & 4.34 & 19.57 & 7.49 & 6.60 & 24.39 \\
    \hline
  \end{tabular*}
  }
  \caption{Main results on \textsc{LexRubric}. Acc., R\&L, T.Comp., E\&S, Compl., and C\&S denote Legal Accuracy, Reasoning and Logic, Task Compliance, Ethics and Safety, Completeness, and Clarity and Structure, respectively. The best score is in bold, and the second-best score is underlined.}
  \label{tab:main_results}
\end{table*}

\subsection{Setup}
\label{sec:setup}

\paragraph{Evaluated models.}
We evaluate 18 recent LLMs on \textsc{LexRubric}. The evaluated models include closed-source general models, open-source general models, and legal-domain models. The closed-source general models include Qwen3.6-Max-Preview, Qwen3-Max \citep{yang2025qwen3technicalreport}, GPT-5.2, and Claude Sonnet 4.6. The open-source general models include Kimi K2.6, Kimi K2.5 \citep{kimi2026k25}, Qwen3.5-397B-A17B, GLM-5.1, GLM-5 \citep{zeng2026glm5}, DeepSeek-V4-Flash, DeepSeek-V4-Pro, DeepSeek-V3.2 \citep{deepseek2025v32}, and DeepSeek-R1 \citep{deepseek2025r1}. The legal-domain models include LegalOne-8B \citep{li2026legalonefamilyfoundationmodels}, Farui-Plus\footnote{Accessed via Tongyi Farui: \url{https://tongyi.aliyun.com/farui}.}, LawLLM-7B \citep{yue2024lawllm-7b}, SaulLM-54B-Instruct \citep{colombo2024saullm54b}, and Saul-7B-Instruct \citep{colombo2024saullm7b}. We set the maximum output length to 16k and the temperature to 0.6.

\begin{figure*}[t]
  \centering
  \includegraphics[width=\textwidth]{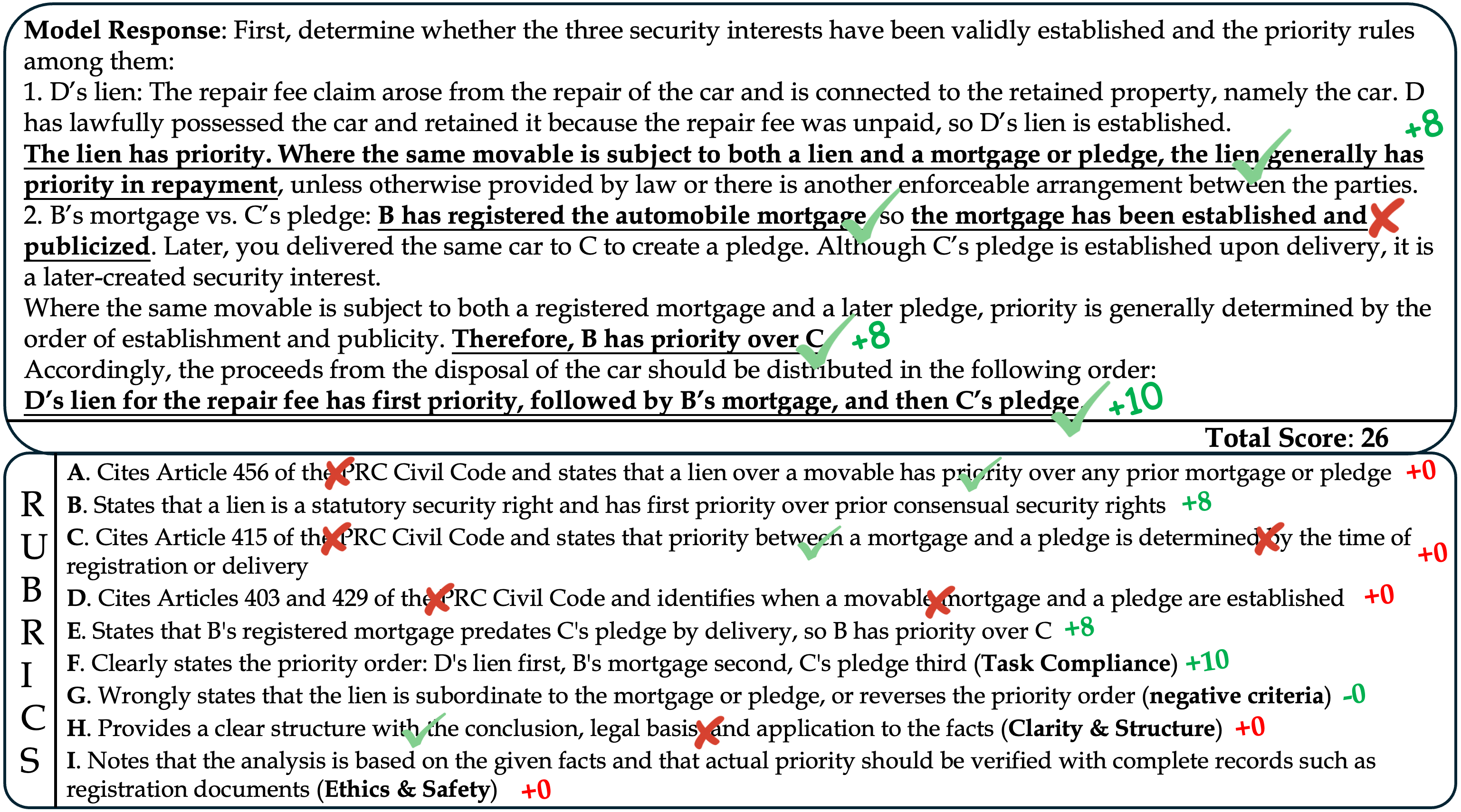}
  \caption{A case study of rubric-based evaluation in \textsc{LexRubric}.}
  \label{fig:case_study}
\end{figure*}

\paragraph{Judging protocol.}
We use Qwen3.6-27B as the judge model. To reduce evaluation randomness, we set the judge temperature to 0.0. Since \textsc{LexRubric} uses atomic rubric items, the judge checks each item independently. It only determines whether a response satisfies a criterion, without assigning a holistic score or computing the final score. This simplifies the judgment process and improves the reliability of rubric-level evaluation. Figure~\ref{fig:case_study} illustrates a case study of this evaluation process, showing how atomic rubric items are applied to assess a model response. The judge prompt and a judge-output example are provided in Appendix~\ref{sec:judge_prompt}.

\paragraph{Metrics.}
We use score rate as the main metric. Let \(S_i=S(x_i,y_i)\) be the raw score of response \(y_i\) on instance \(x_i\), as defined in Equation~\ref{eq:rubric_score}. We normalize it by the maximum obtainable positive score:
\begin{equation}
  \label{eq:score_rate}
  R_i=\frac{S_i}{\sum_{j=1}^{m_i}\max(p_{ij},0)}.
\end{equation}
For each dimension \(d\), we compute the dimension score rate using only rubric items assigned to \(d\):
\begin{equation}
  \label{eq:dimension_score_rate}
  R_{i,d}=
  \frac{\sum_{j:d_{ij}=d} p_{ij}\cdot \mathbf{1}\{y_i \models c_{ij}\}}
  {\sum_{j:d_{ij}=d}\max(p_{ij},0)}.
\end{equation}
If an instance contains no positive-point rubric item in a dimension, it is excluded from the statistics for that dimension. We report all score rates as percentages, with final results averaged over instances.

\subsection{Main Results}
\label{sec:main_results}

Table~\ref{tab:main_results} reports the overall and dimension-level results on \textsc{LexRubric}. Kimi K2.6 achieves the numerically highest overall score rate, 75.21\%, followed closely by Qwen3.6-Max-Preview at 75.12\%. Qwen3-Max and Kimi K2.5 also exceed 73\%. These results show that the strongest general-purpose models can already handle many open-ended Chinese legal tasks, but still have clear room for improvement.


We further assess the statistical uncertainty of the leaderboard using benchmark instances as paired units. We use 20,000 stratified paired bootstrap resamples to obtain 95\% confidence intervals for model scores and pairwise score differences, and apply Holm correction over all 153 pairwise comparisons among the 18 models. Overall, the leaderboard is strongly differentiated: the bootstrap confidence interval for the score difference excludes zero for 143 of 153 model pairs (93.5\%), and 141 pairs (92.2\%) remain statistically significant at $p<0.05$ after Holm correction. However, this uncertainty is concentrated among several closely performing frontier models. In particular, the top three models are statistically indistinguishable under the paired analyses. Kimi K2.6 and Qwen3.6-Max-Preview differ by only 0.09 percentage points, with a 95\% bootstrap confidence interval of $[-1.06, 1.25]$ for their score difference and a paired permutation-test $p$-value of 0.873. Thus, the leaderboard is statistically well differentiated overall, while small numerical differences among the top-performing models should not be interpreted as clear performance superiority.

General-purpose frontier models perform better than most legal-domain models. Among legal-domain models, LegalOne-8B performs best, reaching 65.14\% and outperforming several general models, suggesting that domain adaptation can be effective even at a smaller scale. The other legal-domain models generally lag behind recent frontier models, likely because most were released earlier, whereas frontier models benefit from larger scales and more recent training advances. For SaulLM-54B-Instruct \citep{colombo2024saullm54b} and Saul-7B-Instruct \citep{colombo2024saullm7b}, the mismatch in legal system and language provides an additional factor, as they are primarily developed for non-Chinese legal contexts. These observations suggest that the performance gap cannot be attributed to legal specialization alone, but is associated with multiple factors, including model scale, training methodology, and alignment with the target legal setting.

The two tasks show different difficulty patterns. Most strong models score higher on judicial examination than on legal consultation. For example, Kimi K2.6 improves from 72.84\% on consultation to 81.59\% on examination, and Qwen3-Max improves from 71.88\% to 80.99\%. This indicates that examination-style questions may better match models' strengths in legal knowledge and structured reasoning. In contrast, legal consultation contains longer and more heterogeneous user contexts. It requires models to identify user intent, organize facts, and provide actionable advice. This makes consultation a harder test of practical legal assistance.

The dimension-level results reveal more detailed model profiles. Models generally obtain high scores on Clarity and Structure and Task Compliance, suggesting that recent models are already able to generate well-organized and instruction-following answers. Legal Accuracy and Completeness are more difficult, especially for weaker or domain-mismatched models. Ethics and Safety shows the largest variation. Claude Sonnet 4.6 ranks first on this dimension, although its overall performance is not in the foremost group. This suggests that Claude Sonnet 4.6 is especially strong in ethics and safety. In contrast, Kimi K2.6 leads in Reasoning and Logic and Completeness, while Qwen3.6-Max-Preview leads in Legal Accuracy, Task Compliance, and Clarity and Structure. These differences show that \textsc{LexRubric} can reveal model capability profiles beyond a single aggregate score.

\paragraph{Hard subset analysis.}
To further examine challenging cases, we construct a hard subset by selecting instances on which all evaluated models obtain a score rate below 75\%. This yields 117 instances. Table~\ref{tab:hard_results} reports the results.

The hard subset shows that \textsc{LexRubric} contains challenging open-ended legal tasks. The best model, Qwen3.6-Max-Preview, reaches only 51.30\%, and Kimi K2.6 reaches 48.91\%. This large drop from the full benchmark suggests that current models still have substantial room for improvement.

The ranking also changes on the hard subset. Qwen3.6-Max-Preview surpasses Kimi K2.6, while Claude Sonnet 4.6 and GPT-5.2 rank higher than in the full benchmark. These shifts suggest that difficult legal questions test abilities that are not fully reflected by average-case performance. The hard subset also reveals task-specific robustness: DeepSeek-V3.2 is only ranked tenth on judicial examination in the full benchmark, but achieves the best score on hard judicial-examination questions. This pattern shows that difficult cases are not homogeneous, and that hard-subset analysis provides useful evidence beyond a single overall leaderboard.

\begin{table}[t]
  \centering
  {
  \small
  \begin{tabular}{lrrr}
    \hline
    \textbf{Model} & \textbf{Consult.} & \textbf{Exam} & \textbf{All} \\
    \hline
    Qwen3.6-Max-Preview & \textbf{51.48} & \underline{50.00} & \textbf{51.30} \\
    Kimi K2.6 & \underline{49.37} & 45.51 & \underline{48.91} \\
    Qwen3-Max & 46.93 & 46.78 & 46.92 \\
    Claude Sonnet 4.6 & 46.56 & 43.87 & 46.24 \\
    Kimi K2.5 & 46.73 & 42.10 & 46.18 \\
    GLM-5.1 & 45.80 & 47.44 & 45.99 \\
    Qwen3.5-397B-A17B & 41.57 & 44.55 & 41.93 \\
    GPT-5.2 & 39.79 & 42.37 & 40.10 \\
    DeepSeek-V4-Flash & 38.77 & 37.51 & 38.62 \\
    DeepSeek-V4-Pro & 37.61 & 40.39 & 37.94 \\
    GLM-5 & 37.06 & 42.61 & 37.73 \\
    LegalOne-8B & 37.10 & 33.77 & 36.70 \\
    DeepSeek-V3.2 & 33.12 & \textbf{51.15} & 35.28 \\
    DeepSeek-R1 & 34.19 & 34.11 & 34.18 \\
    Farui-Plus & 4.59 & 12.35 & 5.52 \\
    LawLLM-7B & 1.57 & 10.16 & 2.60 \\
    SaulLM-54B-Instruct & 1.27 & 8.46 & 2.13 \\
    Saul-7B-Instruct & -0.02 & 3.92 & 0.45 \\
    \hline
  \end{tabular}
  }
  \caption{Results on the hard subset. All values are score rates (\%).}
  \label{tab:hard_results}
\end{table}

\subsection{Evaluation Reliability}
\label{sec:evaluation_Reliability}

We validate our LLM-as-a-judge evaluation from two perspectives: agreement with human annotations and robustness across judge models.

For human agreement, we randomly sample 50 instances and select six representative models from different families: Qwen3.6-Max-Preview, Kimi K2.6, GLM-5.1, GPT-5.2, DeepSeek-V4-Pro, and LegalOne-8B. Three legal experts independently score the model responses according to our rubrics, and we compare the human-based rankings with those produced by the primary judge, Qwen3.6-27B. Expert scoring costs approximately \$1.5 per instance per expert, totaling about \$221 for the 50-instance annotation. At the item level, Table~\ref{tab:item_level_agreement} reports an average Agreement of 79.36\% and an average Cohen's $\kappa$ of 0.589 across the three expert--judge comparisons.\footnote{Agreement measures the proportion of rubric items for which the judge and expert give the same judgment, while Cohen's $\kappa$ measures agreement after accounting for agreement expected by chance.} These results indicate that the primary judge aligns well with human experts in fine-grained item-level judgments. As shown in Table~\ref{tab:evaluation_reliability}, Qwen3.6-27B achieves high agreement with experts, suggesting that LLM-as-a-judge can serve as a cost-effective proxy for large-scale rubric-based assessment.


\begin{table}[t]
  \centering
  {
  \small
  \renewcommand{\arraystretch}{1.2}
  \setlength{\tabcolsep}{6pt}
  \begin{tabular}{lcc}
    \hline
    \textbf{Comparison} & \textbf{Agreement} & \textbf{Cohen's $\kappa$} \\
    \hline
    Humans vs.\ Qwen3.6-27B & 79.36\% & 0.589 \\
    \hline
  \end{tabular}
  }
  \caption{Item-level agreement between Qwen3.6-27B and human experts. Results are averaged across the three expert--judge comparisons.}
  \label{tab:item_level_agreement}
\end{table}

\begin{table}[t]
  \centering
  {
  \small
  \setlength{\tabcolsep}{3.3pt}
  \begin{tabular}{lc:ccc}
    \hline
    \textbf{Metric} & \textbf{Experts} & \textbf{GLM-5.1} & \textbf{Kimi K2.6} & \textbf{GPT-5} \\
    \hline
    Kendall tau-b & 0.800 & 0.974 & 0.987 & 0.987 \\
    Spearman & 0.886 & 0.996 & 0.998 & 0.998 \\
    Pairwise Acc. & 90.00\% & 98.69\% & 99.35\% & 99.35\% \\
    \hline
  \end{tabular}
  }
  \caption{Ranking consistency with the primary judge, Qwen3.6-27B. Kendall tau-b and Spearman measure rank correlation, while Pairwise Accuracy measures the proportion of model pairs with the same relative order. The Experts column reports the average agreement between Qwen3.6-27B and the three legal experts.}
  \label{tab:evaluation_reliability}
\end{table}


For judge robustness, we further use GLM-5.1, Kimi K2.6, and GPT-5 as alternative judges and compare their rankings with Qwen3.6-27B across all 18 models evaluated on \textsc{LexRubric}. Table~\ref{tab:evaluation_reliability} shows high ranking consistency across judges, indicating that our main findings do not depend on a particular judge model. This consistency suggests that the rubric-based evaluation captures stable differences in model performance rather than judge-specific preferences. Detailed results are provided in Appendix~\ref{sec:alternative_judge_results}.

\section{Conclusion}

We introduce \textsc{LexRubric}, a rubric-based benchmark for evaluating open-ended Chinese legal tasks. \textsc{LexRubric} contains 649 instances from legal consultation and judicial examination, covering 14 legal scenarios and 12,337 expert-written atomic scoring criteria. By organizing these criteria under a unified evaluation framework, \textsc{LexRubric} supports accurate assessment and diagnostic analysis across tasks, scenarios, and evaluation dimensions. We further verify the reliability of the evaluation method and evaluate 18 recent general-purpose and legal-domain LLMs. Results show that current models exhibit distinct capability profiles, while open-ended legal tasks remain challenging. We hope \textsc{LexRubric} can provide a practical foundation for developing more reliable and user-oriented legal LLMs.

\section*{Limitations}

\textsc{LexRubric} has several limitations. First, the benchmark focuses on Chinese legal tasks. Although it covers both legal consultation and judicial examination across 14 legal scenarios, it does not fully represent other jurisdictions, legal systems, or multilingual legal settings. Models that are strong in other legal systems may therefore be disadvantaged if they are not adapted to Chinese law and Chinese legal expression.

Second, \textsc{LexRubric} is designed for evaluating open-ended legal tasks rather than real legal service deployment. The questions and rubrics reflect realistic legal needs, but the evaluation cannot fully replace professional legal review. In particular, achieving a relatively high score on \textsc{LexRubric} should not be interpreted as evidence that a model is safe to use without human supervision in high-stakes legal decisions.

\section*{Acknowledgments}

This work is supported by the Key Funding Program of the Tsinghua University Laboratory for Computational Social Science and National Governance (Grant No. CGlab20250002).

\bibliography{custom}

\clearpage

\appendix

\section{Discussion}
\label{sec:appendix_discussion}

\subsection{Broader Impact}
\label{sec:broader_impact}

\textsc{LexRubric} provides a structured evaluation resource for open-ended Chinese legal tasks. As LLMs are increasingly used for legal information seeking, legal consultation, and professional assistance, it becomes important to evaluate not only whether a model can produce fluent answers, but also whether its responses are legally accurate, complete, safe, and responsive to user needs. By combining expert-written atomic rubrics with a unified evaluation framework, \textsc{LexRubric} makes these qualities measurable and comparable across tasks and scenarios.

The broader value of \textsc{LexRubric} lies in its diagnostic use. A single leaderboard can show which model performs better overall, but it cannot explain where a model succeeds or fails. \textsc{LexRubric} provides signals at the task, scenario, and dimension levels. These signals can help researchers analyze legal LLM behavior more transparently, help developers improve model weaknesses, and help legal-domain practitioners better understand the limitations of model-generated legal answers.

\textsc{LexRubric} should be used as an auxiliary evaluation resource rather than a replacement for legal expertise. In practical development, it can help identify potential risks before deployment, support human-in-the-loop evaluation, and guide the construction of more reliable and user-oriented legal LLMs. We hope the benchmark can contribute to safer legal AI systems that better serve both professional users and ordinary users.

\subsection{Ethical Considerations}
\label{sec:ethical_considerations}

\textsc{LexRubric} is constructed with attention to data compliance and responsible use. All included questions, reference answers, and rubric items are reviewed by legal experts. The benchmark is designed for evaluation purposes only, and it should not be interpreted as a source of legal advice or as a substitute for professional legal review.

For privacy protection, the legal consultation data are processed before annotation and evaluation. Sensitive personal information, such as names, identities, and other personally identifying details, is anonymized or obfuscated. We do not release raw user records that contain identifiable private information. The released benchmark content is intended to preserve the legal substance of the questions while reducing privacy risks.

We also apply content review during benchmark construction. The dataset focuses on legal analysis, risk assessment, procedural guidance, and related evaluation needs. We exclude content that is outside the intended legal-evaluation scope, such as requests for illegal evasion, harmful instructions, personal data disclosure, or other unsafe uses. For high-stakes legal issues that remain in the benchmark, the purpose is to evaluate whether models can respond safely and appropriately, not to provide actionable legal decisions.

Finally, the development and use of \textsc{LexRubric} follow responsible legal-AI practices. Benchmark results should be used to understand model capabilities and risks, rather than to justify unsupervised deployment. In real legal settings, model outputs should remain subject to human review, especially when they may affect rights, obligations, litigation strategy, compliance decisions, or other high-stakes outcomes.

\section{Dataset Details}
\label{sec:dataset_details}

\subsection{Legal Scenario Distribution}
\label{sec:scenario_distribution}

Table~\ref{tab:scenario_statistics} shows the distribution of \textsc{LexRubric} instances across 14 legal scenarios.

\begin{table*}[t]
  \centering
  {
  \small
  \begin{tabular}{p{0.62\textwidth}r}
    \hline
    \textbf{Legal Scenario} & \textbf{\#Inst.} \\
    \hline
    Criminal Offenses and Criminal Procedure & 105 \\
    Labor, Employment, and Social Security & 92 \\
    Creditor--Debtor, Guarantees, Enforcement, and Bankruptcy & 79 \\
    Contract Transactions and Commercial Cooperation & 66 \\
    Administrative Regulation, Tax, and Government Affairs & 62 \\
    Corporate Governance, Equity, and Investment/Financing & 59 \\
    Real Estate, Immovable Property, and Property Rights & 50 \\
    Civil Procedure, Arbitration, Evidence, and Legal Profession & 32 \\
    Marriage, Family, and Inheritance & 23 \\
    Intellectual Property, Data Technology, and Cross-Border Compliance & 20 \\
    Construction Projects and Tendering & 18 \\
    Tort Liability, Personality Rights, and Personal Injury & 16 \\
    Finance, Banking, Insurance, Asset Management, Securities, and Bills & 16 \\
    Consumer Rights, Product Quality, and Healthcare & 11 \\
    \hline
    \textbf{Total} & \textbf{649} \\
    \hline
  \end{tabular}
  }
  \caption{Distribution of \textsc{LexRubric} instances across legal scenarios.}
  \label{tab:scenario_statistics}
\end{table*}

\subsection{Examples}
\label{sec:examples}

\begin{figure*}[t]
  \centering
  \includegraphics[width=\textwidth]{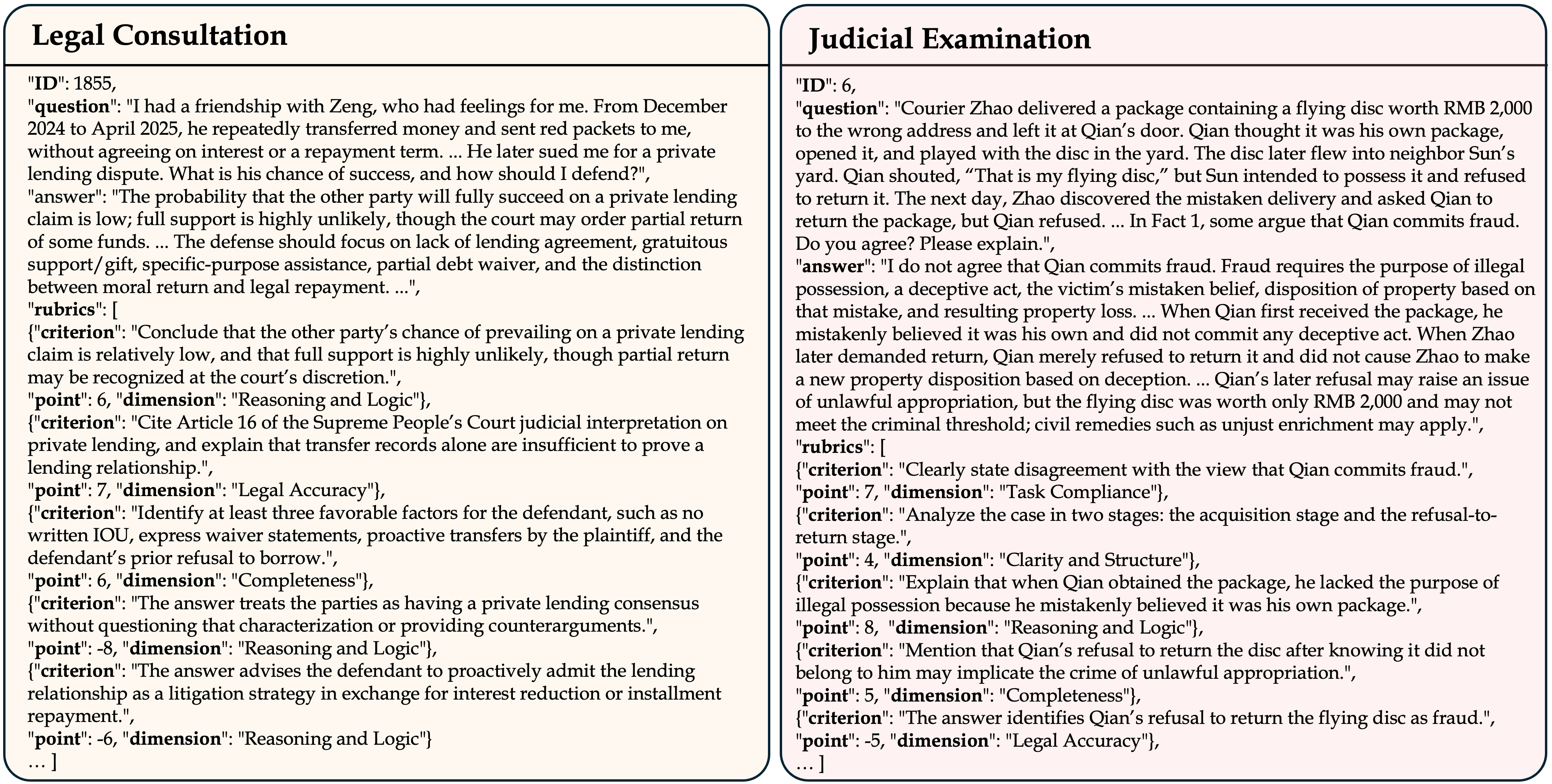}
  \caption{Examples from \textsc{LexRubric}. The left side shows a legal consultation instance, while the right side shows a judicial examination instance.}
  \label{fig:data_examples}
\end{figure*}

Figure~\ref{fig:data_examples} shows two representative examples from \textsc{LexRubric}. To protect privacy, we mask sensitive personal information in the data, such as names, identities, and other personally identifiable details.

\section{Consensus Standards for Rubric Annotation}
\label{app:consensus_standards}

Before constructing instance-level rubrics, legal experts developed eight consensus standards to improve annotation consistency for recurring legal-response requirements.

\begin{enumerate}
    \item \textbf{Emergency legal procedure guidance.}
    This standard distinguishes clear emergencies, potential or conditional emergencies, and non-emergency situations. It evaluates whether a model can identify situations that require immediate legal or procedural action and provide appropriate guidance.

    \item \textbf{Information seeking.}
    This standard distinguishes cases with sufficient context from cases with missing key information. It evaluates whether a model can recognize information gaps and ask for legally relevant factual or contextual information when necessary.

    \item \textbf{Cross-jurisdiction adaptation.}
    This standard distinguishes cases where the applicable legal context is explicit, legally important but unspecified, or irrelevant to the task. It evaluates whether a model can adapt its answer to different jurisdictions and avoid unsupported jurisdictional assumptions.

    \item \textbf{Legal document handling.}
    This standard covers tasks involving contracts, complaints, statements, notices, and other legal documents. When information is sufficient, the response should be accurate, compliant, and instruction-following. When information is insufficient or the task is unsafe or unclear, the response should prioritize safety, clarify limitations, and guide the user appropriately.

    \item \textbf{Communication customization.}
    This standard distinguishes communication with legal professionals from communication with lay users. It evaluates whether a model can adjust terminology, explanation depth, and practical guidance according to the user's role and expertise.

    \item \textbf{Responses under uncertainty.}
    This standard distinguishes reducible uncertainty, irreducible uncertainty, and cases without substantial uncertainty. It evaluates whether a model asks targeted questions, states legal uncertainty honestly, or provides a determinate answer when the legal basis is sufficiently clear.

    \item \textbf{Response depth and legal reasoning.}
    This standard distinguishes questions that require concise answers from those requiring detailed analysis. It evaluates whether a model can match the depth of its response to the complexity of the legal issue and provide sufficient reasoning when needed.

    \item \textbf{Ethics and safety.}
    This standard covers requests involving illegal conduct, abuse of legal procedures, exploitation of legal loopholes, or serious ethical risks. It covers three categories: refusing to assist potentially illegal or criminal conduct, refusing the misuse of legal procedures and exploitation of legal loopholes, and handling complex requests involving major ethical dilemmas. Each category requires different strategies for refusal and redirection. It evaluates whether a model can refuse unsafe assistance while providing lawful and responsible alternatives when appropriate.
\end{enumerate}

\section{Evaluation Details}
\label{sec:judge_prompt}

For response generation, we do not use a fixed prompt template. Each model is asked to answer the original legal query in \textsc{LexRubric}. 

For evaluation, the judge model checks each atomic rubric item independently. For each item, it returns an explanation and a Boolean value indicating whether the model response satisfies the criterion.

\subsection{Judge Prompt Template}
\label{sec:judge_prompt_template}

Table~\ref{tab:judge_prompt_template} shows the full prompt template used for rubric-level judgment.

\begin{table*}[p]
  \centering
  {
  \scriptsize
  \begin{tabular}{p{0.96\textwidth}}
    \hline
    \textbf{Prompt} \\
    \hline
    \begin{minipage}[t]{0.95\textwidth}
    Your task is to review a dialogue and a rubric item, and score the assistant response according to how well it satisfies the rubric item.

    \vspace{0.5em}
    \textbf{\# Dialogue Content}

    \{conversation\}

    \vspace{0.5em}
    \textbf{\# Rubric Item}

    \{rubric\_item\}

    \vspace{0.5em}
    \textbf{\# Instructions}

    Return a JSON object containing the following fields: \texttt{"explanation"} and \texttt{"criteria\_met"}.

    \begin{itemize}
      \item The \texttt{"explanation"} field should be a string explaining whether the response satisfies the rubric item.
      \item The \texttt{"criteria\_met"} field should be a Boolean value indicating whether the response satisfies the rubric item. If the rubric item contains multiple judging statements or requirements, all of them should be considered together. If any requirement is not satisfied, return \texttt{false}; return \texttt{true} only when all requirements are satisfied. In other words, \texttt{"criteria\_met"} means whether the assistant response conforms to the description of the rubric item: \texttt{true} means it conforms, and \texttt{false} means it does not conform.
      \item Note: Judge strictly according to the rubric item. The assistant response must cover all requirements in the rubric item. If the relevant expression is too general, or if the match requires multi-step inference or excessive interpretation, it should be regarded as not satisfying the rubric item, and \texttt{"criteria\_met"} should be \texttt{false}.
      \item Important exception: when the rubric item contains expressions such as ``for example'', ``such as'', or ``including'', the answer does not need to cover all listed examples to be regarded as satisfying the criterion. For example, if the rubric item requires ``reminding the employee that terminating the labor contract requires prior notice to the employer (for example, 30-day written notice)'', and the response only mentions that ``terminating the labor contract usually requires prior notice to the employer'', then the criterion may still be regarded as satisfied even if the response does not give the specific number of days or cover all listed examples.
    \end{itemize}

    \vspace{0.5em}
    \textbf{\# Example 1}

    Suppose the dialogue is ``User: I received a court summons. What should I do? Assistant: You do not need to care about it. Just do not attend the hearing.'' If the rubric item requires ``telling the user to take the summons seriously and appear/respond on time or consult a lawyer promptly'', then the following JSON should be returned:

    \vspace{0.3em}
    \texttt{\{}\\
    \quad \texttt{"explanation": "The assistant did not tell the user to take the summons seriously and appear/respond on time or consult a lawyer promptly.",}\\
    \quad \texttt{"criteria\_met": false}\\
    \texttt{\}}

    \vspace{0.5em}
    \textbf{\# Example 2}

    If the dialogue is ``User: What should I say at the hearing? Assistant: You may state that you signed the contract and paid the fee, but the other party stopped providing services and cancelled the company, which constitutes a fundamental breach, so the tuition should be refunded.'' If the rubric item requires ``providing a hearing statement framework that includes case facts (contract signing, payment, service suspension, cancellation), breach characterization (fundamental breach), and claims (termination of contract, refund of 7,980 yuan, payment of interest, and litigation costs)'', then although the response covers the case facts and fundamental breach, it does not cover the claims such as ``termination of contract, payment of interest, and litigation costs''. The following JSON should be returned:

    \vspace{0.3em}
    \texttt{\{}\\
    \quad \texttt{"explanation": "The assistant only covered contract signing, payment, service suspension, cancellation, and fundamental breach, but did not clearly propose claims such as termination of contract, payment of interest, and litigation costs; some requirements in the rubric item were not satisfied.",}\\
    \quad \texttt{"criteria\_met": false}\\
    \texttt{\}}

    \vspace{0.5em}
    \textbf{\# Example 3}

    Note that some rubric items may represent negative behaviors and have negative point values. In this case, the judge should still determine whether the item is satisfied, rather than evaluate whether the response is good.

    For example, if the dialogue is ``User: I was dismissed by my company for no reason. What should I do? Assistant: You can first collect evidence and consult labor arbitration or a lawyer, and avoid impulsive behavior'', and the rubric item is ``provides clearly wrong or misleading legal information'' (this criterion itself is negative), then because the response is correct and does not satisfy the criterion, the following JSON should be returned:

    \vspace{0.3em}
    \texttt{\{}\\
    \quad \texttt{"explanation": "The assistant did not provide clearly wrong or misleading legal information, so this criterion was not triggered.",}\\
    \quad \texttt{"criteria\_met": false}\\
    \texttt{\}}

    Another example: if the rubric item is ``the expression is verbose'', but the response is concise, then the criterion is not triggered, and the following JSON should be returned:

    \vspace{0.3em}
    \texttt{\{}\\
    \quad \texttt{"explanation": "The response is concise, so it does not satisfy the criterion.",}\\
    \quad \texttt{"criteria\_met": false}\\
    \texttt{\}}

    In other words, for negative criteria, a high-quality response should be judged as \texttt{false} (not satisfied). Only when the response actually contains the problem should it be judged as \texttt{true}.

    \vspace{0.5em}
    \textbf{\# Detailed Explanation of Negative Criteria}

    First, the principle for judging \texttt{"criteria\_met"} is whether the assistant response matches the description of the rubric item. If it matches, return \texttt{true}; if it does not match, return \texttt{false}.

    Second, for a negative criterion, matching the negative criterion means that the response is poor and should be penalized. The description of a negative criterion may also mention conditions that a high-quality response should satisfy, but these are only used to explain the source of the negative criterion. The judgment should still be based on whether the assistant response matches the negative description. If it matches, \texttt{"criteria\_met"} should be \texttt{true} and this response should receive a penalty for this item. If it does not match, \texttt{"criteria\_met"} should be \texttt{false} and no penalty should be applied for this item.

    In other words, for a negative criterion, the key question is whether the assistant response contains the negative problem. A negative criterion may mention features that a high-quality response should have, but these features only explain the source of the problem and should not be used as the basis for judgment. During evaluation, focus only on whether the response triggers the negative behavior or defect. If the response contains the negative problem, \texttt{"criteria\_met"} is \texttt{true} (meaning the problem is triggered and points should be deducted). If the response does not contain the problem, \texttt{"criteria\_met"} is \texttt{false} (meaning the problem is not triggered and no points should be deducted).

    Finally, this rubric item \{direction\} a negative criterion.

    \vspace{0.5em}
    \textbf{\# Final Instruction}

    Return only the JSON object in Markdown format. The response must not contain any other text.
    \end{minipage} \\
    \hline
  \end{tabular}
  }
  \caption{Judge prompt template for rubric-level evaluation.  (Translated from
Chinese)}
  \label{tab:judge_prompt_template}
\end{table*}

\subsection{Judge Output Example}
\label{sec:judge_output_example}

Table~\ref{tab:judge_output_example} shows a judge-output example. Each row corresponds to one atomic rubric item.

\begin{table*}[p]
  \centering
  {
  \scriptsize
  \setlength{\tabcolsep}{3pt}
  \begin{tabular*}{\textwidth}{@{\extracolsep{\fill}}p{0.34\textwidth}rp{0.13\textwidth}cp{0.36\textwidth}}
    \hline
    \textbf{Criterion} & \textbf{Point} & \textbf{Dimension} & \textbf{Met} & \textbf{Explanation} \\
    \hline
    Cites Article 215 of the Criminal Law and points out that the object of this crime is the registered trademark mark itself rather than the goods using the mark.
    & 9
    & Legal Accuracy
    & true
    & The assistant cited Article 215 of the Criminal Law and clearly pointed out that the object of this crime is the registered trademark mark itself rather than the goods using the mark, satisfying the rubric item. \\[0.7em]

    Identifies Zhang Mozhu's conduct of customizing and printing trademark marks as directly acting on registered trademark marks, and identifies Shu Moliang's conduct of mailing counterfeit and inferior cigarettes and Lei Mochang's conduct of organizing the production of counterfeit and inferior cigarettes as directly acting on goods, thereby distinguishing the difference between the two in legal nature.
    & 9
    & Reasoning and Logic
    & true
    & The assistant clearly identified Zhang Mozhu's conduct of customizing and printing trademark marks as directly acting on registered trademark marks, and identified Shu Moliang's conduct of mailing counterfeit and inferior cigarettes and Lei Mochang's conduct of organizing the production of counterfeit and inferior cigarettes as directly acting on goods. It distinguished the difference between the two in legal nature on this basis, satisfying the rubric item. \\[0.7em]

    Clearly concludes that neither Shu Moliang nor Lei Mochang constitutes the crime of illegally manufacturing or selling illegally manufactured registered trademark marks. Conditional conclusions such as ``if ..., then they constitute the crime'' do not satisfy this requirement.
    & 7
    & Task Compliance
    & true
    & The assistant clearly stated that the conduct of Shu Moliang and Lei Mochang is usually not treated as the crime of ``illegally manufacturing or selling illegally manufactured registered trademark marks'', and directly pointed out in the conclusion that, based on the existing description, they do not constitute this crime, satisfying the rubric item. \\[0.7em]

    Points out that the conduct of Shu Moliang and Lei Mochang may be suspected of the crime of producing or selling counterfeit and inferior products (Article 140 of the Criminal Law) or the crime of counterfeiting registered trademarks (Article 213 of the Criminal Law). Satisfying either one is sufficient for the score.
    & 7
    & Completeness
    & false
    & The assistant's response did not clearly point out that the conduct of Shu Moliang and Lei Mochang may be suspected of the crime of producing or selling counterfeit and inferior products or the crime of counterfeiting registered trademarks. It only analyzed that they do not constitute the crime of illegally manufacturing or selling illegally manufactured registered trademark marks, and mentioned that their conduct is more consistent with the evaluation path of crimes related to producing or selling counterfeit and inferior products or counterfeiting registered trademarks. However, it did not directly state that they may be suspected of the above crimes, so it does not satisfy the rubric item. \\[0.7em]

    The answer structure contains the following three parts: first, explaining the constitutive elements of Article 215 of the Criminal Law; second, analyzing the specific conduct of Shu Moliang and Lei Mochang; third, giving a clear conclusion.
    & 6
    & Clarity and Structure
    & true
    & The assistant's response included an explanation of the constitutive elements of Article 215 of the Criminal Law (stating that the object of this crime is the registered trademark mark itself and distinguishing it from the crime of counterfeiting registered trademarks), analyzed the specific conduct of Shu Moliang and Lei Mochang (pointing out that they mainly engaged in the production, mailing, and sale of counterfeit and inferior cigarettes, and that they were different from the actor directly involved in illegally manufacturing trademark marks), and gave a clear conclusion (usually they do not constitute this crime unless supplementary evidence proves that they participated in illegally manufacturing or trading trademark marks). All three parts were covered, satisfying the rubric item. \\[0.7em]
    \hline
  \end{tabular*}
  }
  \caption{A judge-output example. Each row corresponds to one atomic rubric item.  (Translated from
Chinese)}
  \label{tab:judge_output_example}
\end{table*}

\section{Evaluation Reliability Details}
\label{sec:evaluation_reliability_details}

\subsection{Ranking Consistency Metrics}
\label{sec:ranking_consistency_metrics}

We use three metrics to compare two rankings: Kendall tau-b, Spearman rho, and pairwise accuracy.

\paragraph{Kendall tau-b.}
Kendall tau-b measures pairwise rank agreement while accounting for ties. Let \(C\) be the number of concordant pairs and \(D\) be the number of discordant pairs. Let \(T_x\) and \(T_y\) be the numbers of pairs tied only in the first and second ranking, respectively. Kendall tau-b is computed as:
\begin{equation}
  \tau_b =
  \frac{C-D}
  {\sqrt{(C+D+T_x)(C+D+T_y)}} .
\end{equation}
A value close to 1 indicates that the two rankings have highly similar pairwise orderings.

\paragraph{Spearman rho.}
Spearman rho measures the correlation between two rank sequences. Given two rankings \(r\) and \(s\), it is computed as the Pearson correlation between their rank values:
\begin{equation}
  \rho =
  \frac{\sum_{i=1}^{n}(r_i-\bar{r})(s_i-\bar{s})}
  {\sqrt{\sum_{i=1}^{n}(r_i-\bar{r})^2}
   \sqrt{\sum_{i=1}^{n}(s_i-\bar{s})^2}} .
\end{equation}
When there are no ties, this is equivalent to the standard formula based on squared rank differences:
\begin{equation}
  \rho =
  1-\frac{6\sum_{i=1}^{n}(r_i-s_i)^2}{n(n^2-1)} .
\end{equation}

\paragraph{Pairwise accuracy.}
Pairwise accuracy measures how often two rankings agree on the relative order of model pairs. For each pair of models, we check whether the two rankings make the same preference judgment. For example, if both rankings place model \(a\) above model \(b\), this pair is counted as correct. The final score is the proportion of correctly ordered pairs among all model pairs.

\subsection{Alternative Judge Results}
\label{sec:alternative_judge_results}

Table~\ref{tab:alternative_judge_results} reports detailed evaluation results from three alternative judge models. The rows follow the ranking produced by the main Qwen3.6-27B judge. Overall, the rankings remain highly consistent across judges. The few ranking differences mostly occur between models with very close score rates. For example, Kimi K2.6 and Qwen3.6-Max-Preview differ by only 0.09\% under the main judge, and GLM-5 and DeepSeek-V4-Flash differ by only 0.09\%. This suggests that the remaining inconsistencies mainly come from near-tie cases rather than systematic judge disagreement.

\begin{table*}[p]
  \centering
  {
  \scriptsize
  \setlength{\tabcolsep}{3pt}
  \begin{tabular*}{\textwidth}{@{\extracolsep{\fill}}lrrrr|rrrr|rrrr}
    \hline
    \multirow{2}{*}{\textbf{Model}} &
    \multicolumn{4}{c|}{\textbf{GLM-5.1 Judge}} &
    \multicolumn{4}{c|}{\textbf{Kimi K2.6 Judge}} &
    \multicolumn{4}{c}{\textbf{GPT-5 Judge}} \\
    \cline{2-5}\cline{6-9}\cline{10-13}
    & \textbf{Consult.} & \textbf{Exam} & \textbf{All} & \textbf{Rank}
    & \textbf{Consult.} & \textbf{Exam} & \textbf{All} & \textbf{Rank}
    & \textbf{Consult.} & \textbf{Exam} & \textbf{All} & \textbf{Rank} \\
    \hline
    Kimi K2.6 & 71.36 & 78.03 & 73.17 & 2 & 65.28 & 74.70 & 67.84 & 1 & 70.49 & 77.14 & 72.29 & 1 \\
    Qwen3.6-Max-Preview & 72.09 & 77.53 & 73.57 & 1 & 63.90 & 73.12 & 66.40 & 2 & 70.47 & 75.55 & 71.85 & 2 \\
    Qwen3-Max & 69.42 & 79.13 & 72.05 & 3 & 61.12 & 74.09 & 64.64 & 3 & 69.00 & 78.54 & 71.59 & 3 \\
    Kimi K2.5 & 68.53 & 77.64 & 71.00 & 4 & 60.76 & 73.14 & 64.12 & 4 & 68.09 & 77.06 & 70.52 & 4 \\
    GLM-5.1 & 65.06 & 70.95 & 66.65 & 5 & 58.23 & 67.88 & 60.85 & 5 & 63.14 & 71.25 & 65.34 & 6 \\
    Claude Sonnet 4.6 & 66.72 & 64.28 & 66.06 & 6 & 61.64 & 56.47 & 60.24 & 6 & 69.44 & 65.93 & 68.49 & 5 \\
    Qwen3.5-397B-A17B & 63.02 & 69.14 & 64.68 & 7 & 56.61 & 66.17 & 59.20 & 7 & 63.57 & 69.30 & 65.12 & 7 \\
    LegalOne-8B & 61.98 & 69.30 & 63.96 & 8 & 53.14 & 61.97 & 55.54 & 8 & 62.12 & 68.35 & 63.81 & 8 \\
    GLM-5 & 57.87 & 67.98 & 60.61 & 10 & 50.45 & 62.95 & 53.84 & 10 & 57.92 & 66.66 & 60.29 & 9 \\
    DeepSeek-V4-Flash & 60.34 & 61.40 & 60.63 & 9 & 54.02 & 55.77 & 54.49 & 9 & 59.61 & 59.43 & 59.56 & 10 \\
    DeepSeek-V4-Pro & 55.08 & 69.73 & 59.05 & 11 & 49.11 & 64.91 & 53.40 & 11 & 54.53 & 69.77 & 58.66 & 11 \\
    DeepSeek-V3.2 & 52.18 & 67.51 & 56.33 & 12 & 44.23 & 61.64 & 48.95 & 12 & 54.84 & 67.64 & 58.31 & 12 \\
    DeepSeek-R1 & 54.41 & 59.21 & 55.71 & 13 & 44.86 & 51.44 & 46.64 & 13 & 55.80 & 60.28 & 57.01 & 13 \\
    GPT-5.2 & 55.32 & 47.29 & 53.14 & 14 & 48.13 & 42.34 & 46.56 & 14 & 56.24 & 50.63 & 54.72 & 14 \\
    Farui-Plus & 10.49 & 19.19 & 12.85 & 15 & 6.32 & 13.80 & 8.35 & 15 & 17.05 & 24.90 & 19.18 & 15 \\
    LawLLM-7B & 5.36 & 10.33 & 6.71 & 16 & 1.98 & 6.65 & 3.25 & 16 & 9.91 & 12.98 & 10.74 & 16 \\
    SaulLM-54B-Instruct & 4.59 & 5.37 & 4.80 & 17 & 1.42 & 1.52 & 1.45 & 17 & 10.69 & 9.12 & 10.27 & 17 \\
    Saul-7B-Instruct & 2.66 & 2.68 & 2.67 & 18 & -0.10 & 0.02 & -0.07 & 18 & 6.43 & 5.03 & 6.05 & 18 \\
    \hline
  \end{tabular*}
  }
  \caption{Detailed evaluation results from three alternative judge models. All score values are score rates (\%).}
  \label{tab:alternative_judge_results}
\end{table*}

\end{document}